\documentclass[journal]{IEEEtran}

\usepackage{cite}
\usepackage{amsmath,amssymb,amsfonts}
\usepackage{algorithm}
\usepackage{algorithmic}
\usepackage{graphicx}
\usepackage{textcomp}
\usepackage{hyperref}

\begin{document}

\title{Application of Genetic Algorithms to the Multiple Team Formation Problem}

\author{Jos{\'e} G. M. Esgario,
Iago E. da Silva, and
Renato A. Krohling
\thanks{J. G. M. Esgario is with Graduate Program in Computer Science, Federal University of Esp\'irito Santo, Brazil (e-mail: guilherme.esgario@gmail.com).}
\thanks{I. E. da Silva is with the Department of Production Engineering, Federal University of Esp\'irito Santo, Brazil (e-mail: iagoegias@gmail.com).}
\thanks{R. A. Krohling is with the Department of Production Engineering and Graduate Program in Computer Science, Federal University of Esp\'irito Santo, Brazil (e-mail: krohling.renato@gmail.com).}
}



\maketitle

\begin{abstract}
Allocating of people in multiple projects is an important issue considering the efficiency of groups from the point of view of social interaction. In this paper, based on previous works, the Multiple Team Formation Problem (MTFP) based on sociometric techniques is formulated as an optimization problem taking into account the social interaction among team members. To solve the resulting optimization problem we propose a Genetic Algorithm due to the NP-hard nature of the problem. The social cohesion is an important issue that directly impacts the productivity of the work environment. So, maintaining an appropriate level of cohesion keeps a group together, which will bring positive impacts on the results of a project. The aim of the proposal is to ensure the best possible effectiveness from the point of view of social interaction. In this way, the presented algorithm serves as a decision-making tool for managers to build teams of people in multiple projects. In order to analyze the performance of the proposed method, computational experiments with benchmarks were performed and compared with the exhaustive method. The results are promising and show that the algorithm generally obtains near-optimal results within a short computational time.
\end{abstract}


\IEEEpeerreviewmaketitle

\section{Introduction}

\IEEEPARstart{E}{empirical} studies in the area of multi-project management are still rare \cite{PATANAKUL2009216}, and usually when building a team of people, technical skills are the most considered attributes by project managers. However, recent studies have shown that either success or failure of a group often depends on: a) the interdependence of group skills; b) integration; c) trust; and d) the technical skills of each member \cite{BAIDEN2011129}. In addition, an appropriate level of cohesion is required to get people together and make them collaborate to build the basis of a quality work \cite{Hoegl2001}. In sociological theory, some studies point out that team cohesion is an important variable in the emergence of consensus among team members and that cohesion is directly responsible for positively impacting the productivity of a work environment, influencing team motivation, morale, coordination of efforts, productivity, satisfaction and cooperation \cite{DWIVEDULA2010}. Besides that, Social Identity Theory suggests that the more a person identifies himself as belonging to a group, the more that person will actively contribute to achieve a common goal \cite{MAURER2010629}.

Aiming to determine the effects of social cohesion on a group of people, multiple tasks were performed by 50 army teams and a positive effect of social cohesion on the teams' physical and mental performance was identified \cite{jordan2002}. Seers, Petty and Cashman \cite{seers1995} reported a study of 103 manufacturing workers that found a positive association between job satisfaction, motivation, and group cohesion. Sanders and Nauta \cite{sanders2004} pointed out that how increased cohesion reduces employee absenteeism.

It is therefore obvious that social interaction plays an important role in achieving the success of a project. However, only in 2009 the first attempt was made by Lappas, Liu and Terzi \cite{lappas2009} to allocate people with different skills to a group while seeking to maximize their social compatibilities. Since then, several other approaches to the problem have been developed and other works have been published proposing better models for the determination of an optimal solution with lower computational costs. For instance, it was proposed a new allocation model \cite{Fathian2017} that takes into account the collaboration among members and seeks to minimize the likelihood of someone leaving the project which would decrease performance and put the project at risk. In \cite{GUTIERREZ2016} the problem of allocating multiple people (either full-time or in smaller time fractions) to various groups is called the Multiple Team Formation Problem (MTFP) and some algorithms were proposed for the solution of the problem. The MTFP has received attention in the scientific literature \cite{campion1993,FITZ2005}, however the majority of these works are based only on the psychological and behavioral perspective.

Ballesteros-P\'erez, Gonz\'alez-Cruz and Fern\'andez-Diego \cite{BALLES2012} developed a method for allocating people in multiple projects, so that the combination of human resources allocated to different working groups, maximizes the efficiency of groups from the point of view of social interaction. This approach allocates individuals based on sociometric techniques whereas the main point was to present a manual calculation method that determines a sufficiently good allocation, in a very practical way - the optimal solution is not guaranteed. Although such a method is quite practical for a small number of people, since the NP-Hard nature of the problem the method becomes prohibitive when this number increases. For computer-based calculations, where the optimal solution to the problem at hand is guaranteed,  the authors have limited themselves to describing that a computational application must take into account the constraints and information given by the project managers. Unfortunatelly, the computation of all possible permutations for determining the optimall allocations is viable only for teams with small number of individuals.

Yannibelli and Amandi \cite{YANNI2012} proposed a deterministic crowding evolutionary algorithm for the formation of collaborative learning teams, so that the roles of students in a group are balanced. The idea of applying an evolutionary algorithm to solve this problem aims to find good solutions in a short time period. Other works also made use of evolutionary approaches for grouping people based on their interpersonal relationships \cite{lin2008,agustin2011,chen2012,chen2013}. Da Silva and Krohling \cite{silva2018} presents an algorithm based on Sociometry that incorporates fuzzy numbers to the MTFP and allows the expression of personal preferences provided to the sociometric test in a more natural way.

In this paper, the MTFP is formulated as a cohesion maximization problem subject to constraints of the requirements matrix. To solve the problem, a Genetic Algorithm (GA) is proposed. This paper extends the work developed by Ballesteros-P\'erez, Gonz\'alez-Cruz and Fern\'andez-Diego \cite{BALLES2012} and was inspired by Yannibelli and Amandi \cite{YANNI2012}, who proposed an evolutionary approach to solve the MTFP, which allows finding optimal or semi-optimal solutions to the problem when the number of individuals to be allocated is large and the computation of all possible solutions is not viable. The remainder of this paper is organized as follows: section 2 presents the problem formulation; section 3 is devoted to describing the proposed approach; section 4 presents the experimental results and finally a conclusion is presented in section 5.

\section{Problem Formulation}

\subsection{Group Cohesion and Sociometric Matrix}

\begin{figure}[htbp]
\centerline{\includegraphics[width=3in]{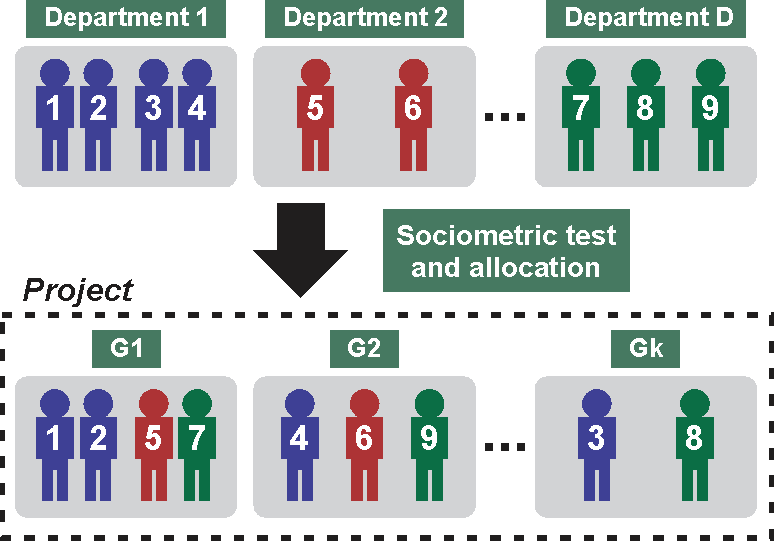}}
\caption{Multiple team formation problem scheme.}
\label{figmtfp}
\end{figure}

Given an organization that develops projects of any kind, which is composed of different individuals belonging to different departments and having different skills, it is desired to allocate each one of the individuals to some group of the organization to carry out a project in a way that maximizes the cohesion as a whole. Fig. \ref{figmtfp} shows a scheme of how MTFP works.

Group cohesion is defined as the degree in which individuals feel accepted or rejected by a given group \cite{BEAL2003}. In developing the proposed algorithm, it was postulated, as in \cite{BALLES2012}, that the result achieved by a team depends greatly on the way in which individuals develop their relationships and interactions. Thus, if it is possible to maximize the cohesion of several groups by bringing together certain people, the whole project is expected to have an optimal performance.

The cohesion degree of a given group can be obtained through several methods, such as: sociometric tests and work environment studies. The sociometric test is a tool used by Sociometry created by Jacob Levy Moreno \cite{moreno1961} to understand how relationships between members of a given group are structured.

In the sociometric test each member accepts or rejects each of the other members of the project. The results of the sociometric test can be obtained through the following questions: 1) ``Would you like to work with which employees?'' \ 2) ``Which employees would you not like to work with?''.

From the results obtained in the sociometric test a matrix called Sociometric Matrix is then constructed, whose values +1, 0 and -1, mean that a given member ``chose'', omitted his opinion or ``rejected'' another member, respectively. The values contained in the main diagonal of the Sociometric Matrix will always be equal to zero since self-evaluations are not allowed.

\subsection{The Sociometric and the Project Requirement Matrices}

To solve the MTFP two matrices are required as input, the Sociometric Matrix $S$ and the Project Requirement Matrix $R$. The Sociometric Matrix $S$ is described by

\begin{equation}
S = 
\bordermatrix{
    & I_1 & \cdots & I_{n_i} \cr
    I_1 & x_{11} & \ldots & x_{1n_i} \cr
    \vdots & \vdots & \ddots & \vdots \cr
    I_{n_i} & x_{n_i1} & \ldots & x_{n_in_i} \cr
    }
\end{equation} where each project member $I_i$ assigns a $x$ value to another member that will be part of the main project to be formed. As stated earlier, $x$ will assume values equal to $-1$, $0$ or $1$ depending on the answers given to sociometric test questions.
Each member $I_i$ is also characterized from their skills, training or in terms of information about which department of the company this member belongs to. For example, $I_7$ is the worker number $7$ in the finance department of a given company, it may be denoted $I_7$ as F7, where the letter ``F'' means ``Department of Finance''.

The other matrix needed to solve the problem is called the Project Requirement Matrix $R$, described by

\begin{equation}\label{eq:reqmatrix}
R = 
\bordermatrix{
    & G_1 & \cdots & G_k \cr
    D_1 & y_{11} & \ldots & y_{1k} \cr
    \vdots & \vdots & \ddots & \vdots \cr
    D_j & y_{j1} & \ldots & y_{jk} \cr
    }
\end{equation} where a certain number of people $y$ from a given department or a group of people with a certain skill are required for each group or subproject $G_l (l = 1, 2, ..., k)$.

\subsection{Problem Constraints and Solution Space}

Once the input matrices are obtained, we proceed to the second phase of the problem. In this phase all Allocation Matrices $A$ are generated, which correspond to the possible solutions of the problem. An Allocation Matrix $A$ is described by
\begin{equation}\label{eq:alocmatrix}
A = 
\bordermatrix{
    & G_1 & \cdots & G_k \cr
    I_1 & a_{11} & \ldots & a_{1k} \cr
    \vdots & \vdots & \ddots & \vdots \cr
    I_{n_i} & a_{n_i1} & \ldots & a_{n_ik} \cr
    }
\end{equation} where the elements $a_{ij} \in \{0,1\} $ indicate to which group or subproject $G_l$ the worker $I_i$ will be allocated. A feasible solutions must necessarily meet the following constraints:

$1)$ Comply with the Project Requirement Matrix \eqref{eq:reqmatrix}.

$2)$ Allocate each individual to only one group according to the equation

\begin{equation}\label{eq:restricao2}
    \sum_{j=1}^k{a_{ij} = 1}, \ \forall i \in 1,2,\dots,n_i
\end{equation}

\subsection{General Cohesion}

To determine which allocation is the one that maximizes the General Cohesion $E_g$ it is necessary compute the General Cohesion of each solution $A$ generated. $E_g$ is calculated as

\begin{equation}\label{eq:gencoe}
    E_g = \sum_{k=1}^{n_k}{W_kE_k}
\end{equation} where $n_k$ is the total number of groups that forms the main project, $W_k$ is the weight of each group given by $W_k = \frac{n_{ik}}{n_i}$, $n_{ik}$ is the number of individuals of a group $k$, $n_i$ is the total number of individuals in the main project and $E_k$ represents the cohesion of the $k$-th group and is defined as

\begin{equation}
    E_k = \frac{\sum_{j=1}^{n_i}{a_{ik}a_{jk}x_{ij}}}{n_{ik}}, \ \forall i \in 1,2,\dots,n_i
\end{equation}

The higher $E_k$ is, the greater is the cohesion among the members of a group.

\section{Proposed Approach}

Genetic Algorithm (GA) originally developed by Holland \cite{holland1975} is a powerful optimization algorithm inspired by the concepts of Darwin' s Theory of Evolution. In this paper, we propose to solve the MTFP using a GA.

The main components of the algorithm are: genetic representation, population initialization, genetic operators (selection, crossover and mutation) and the fitness function. All components are detailed in the following.

\subsection{Genetic Representations}

The individual of the population are possible solutions of the optimization problem, and codified individuals are known as chromosomes. Each chromosome is represented as a vector of dimension $n_i$ where each position $i$ ($i = 1,...,n_i$) of this vector represents an individual of the MTFP. The value of the elements that constitutes these vectors indicates the group in which the $i$-th individual must be allocated. The variables of a chromosome, called genes, use binary coding. The number of bits used in these variables were defined as being equal to the number of groups $n_g$ of the MTFP. Just one bit receive the value 1 and its position indicates the group that the individual will compose.

\begin{figure}[htbp]
\centerline{\includegraphics[width=3.5in]{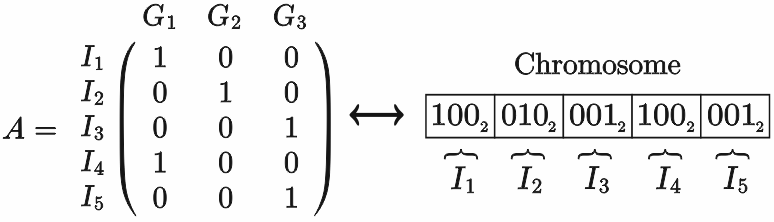}}
\caption{Codification of a Solution Matrix}
\label{figchromosome}
\end{figure}

Fig. \ref{figchromosome} illustrates the coding of a Solution Matrix, taking as example a matrix $A$, containing five individuals and three groups. The chromosome has five genes where each gene is codified by three bits.

The generation of the initial population consisted in the creation of a set of random solutions satisfying the equation \eqref{eq:restricao2}. Initially, genes of the chromosomes are initialized with zeros and for each gene a random bit is selected which will be assigned the value 1.

\subsection{Genetic Operators}

Next, we describe the three genetic operators.

\subsubsection{Selection}

The selection process used by GA is based on Darwin's theory of natural selection, whereas fit individuals are more likely to survive. For the selection of the most fit individuals, the tournament selection method was used since it is a simple and widely used in the literature \cite{miller1995}. Two possible solutions or individuals are selected randomly from the current population, the fitness values of these individuals are compared and the one with the best value is selected for the next generation.

\subsubsection{Crossover}

\begin{figure}[htbp]
\centerline{\includegraphics[width=2.2in]{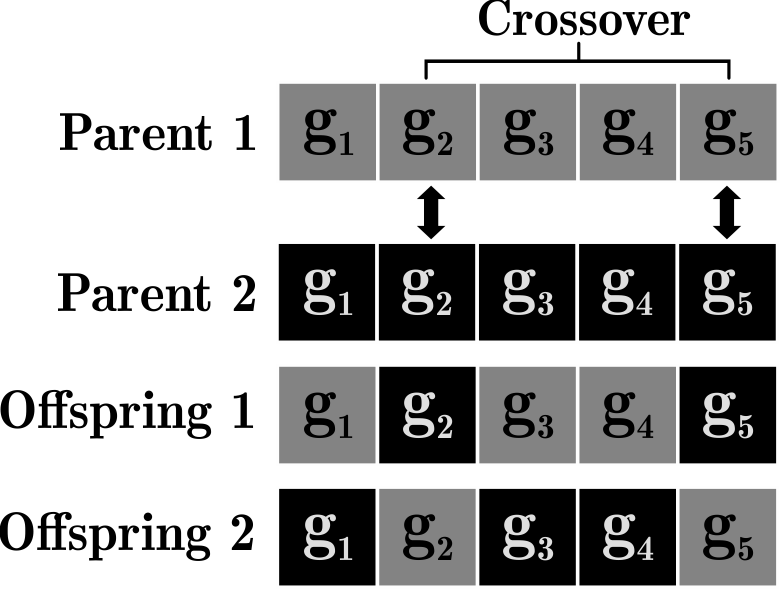}}
\caption{Example of a crossover operator application.}
\label{figcrossover}
\end{figure}

The crossover operator acts on the individuals resulting from the selection process through the exchange of genetic material, thus generating new individuals. Individuals of the population are selected in pairs, called parents, and are crossed so that each gene has a $\beta$ probability of being exchanged between them. The crossing process results in a new population composed of these new pairs of individuals, called offspring. Fig. \ref{figcrossover} shows an example of applying the crossover operator on a pair of individuals. The chromosomes have five genes, the $g_2$ and $g_5$ genes of the parents were exchanged generating two new individuals.

\subsubsection{Mutation}

The mutation operator increases the diversity of the solutions and helps the algorithm to escape from local minima by changing one or more genes from a chromosome by random values. The mutation is performed with a given probability $\alpha$ called mutation probability. For each gene of a chromosome a random value $r$ in the interval $[0,1]$ is generated, if $r < \alpha$ then the mutation is performed. The mutation process employed consists of assigning zero to the gene value and randomly a new bit is selected to receive the value $1$.

\subsection{Fitness function}
Due to the intrinsic constraints of the problem, there is a need to use some constraint-handling strategy that forces the algorithm to produce feasible solutions. Among the several approaches developed to treat constrained optimization problems, many use the penalty method \cite{yi2005} that adds a penalty component to the objective function by transforming the constrained optimization problem into an unconstrained one. The fitness function with the penalty method for the MTFP is calculated according to \begin{equation}
\label{eq:penaltyfunc}
\max \ F(X) = E_g(X) - P(R'(X),R)
\end{equation} $E_g(X)$ represents the General Cohesion calculated by \eqref{eq:gencoe}. $P(R'(X),R)$ is the penalty function that measures how much a possible solution violates the constraints of the problem, i.e., the sum of the module of the difference between the obtained Requirement Matrix $R'(X)$ and the desired Requirement Matrix $R$. The penalty $P(R'(X),R)$ is given by

\begin{equation}
    P(R'(X),R) =  \sum_{j=1}^{n_j}{\sum_{k=1}^{n_k}{ |y'_{jk} - y_{jk}| }}
\end{equation}

Given a Solution Matrix $X$ generated from an individual of the population. The obtained Requirement Matrix is calculated as follows

\begin{equation}
R'(X) = 
\bordermatrix{
    & G_1 & \cdots & G_k \cr
    D_1 & y'_{11} & \ldots & y'_{1k} \cr
    \vdots & \vdots & \ddots & \vdots \cr
    D_j & y'_{j1} & \ldots & y'_{jk} \cr
    }
\end{equation} where the element $y_{jk}'$ of the matrix $R'$ is equal to the sum of individuals of the same department who were allocated to work in the same group and it is calculated as \begin{equation}
    y'_{jk} = \sum_{i=1}^{n_i}{x_{ik}\delta}
\end{equation}

\begin{equation}\label{eq:delta}
\begin{split}
\delta
=\begin{cases}
  1, & D(I_i) = D_j \\
  0, & \textrm{otherwise} \\
  \end{cases} 
\end{split}
\end{equation} where $D(I_i)$ represents the department of the $i$-th individual.

The pseudo-code of the approach used to solve the MTFP proposed, is described in Algorithm \ref{alg:ga}.

\begin{algorithm}[H]
    \caption{Genetic Algorithm}
    \label{alg:ga}
    \hspace*{\algorithmicindent} \textbf{Input:} $R$: Requirement Matrix, $S$: Sociometric Matrix \\
 	\hspace*{\algorithmicindent} \textbf{Output:} $A$: Solution Matrix 
    \begin{algorithmic}[1]
    	\STATE $P \gets$ Initialize population
    	\WHILE{number of generation is not met}
    	    \STATE $F \gets$ Compute fitness($P$, $R$, $S$)
    		\STATE $P \gets$ Selection($P$, $F$)
    		\STATE $P \gets$ Crossover($P$, $\beta$)
    		\STATE $P \gets$ Mutation($P$, $\alpha$)
    	\ENDWHILE
    	\STATE $A \gets$ Solution Matrix($P_{best}$)
    \end{algorithmic}    
\end{algorithm}

\section{Experimental Results}

\subsection{Benchmarks}

In order to perform the experiments, seven datasets were developed. The features of each dataset are presented in the table \ref{tabdataset}.

\begin{table}[htbp]
\caption{Features of the proposed datasets}
\begin{center}
\begin{tabular}{cccc}
\hline
\textbf{\textit{Dataset}} & \textbf{\textit{Individuals}} & \textbf{\textit{Groups}} & \textbf{\textit{Departments}} \\
\hline
1 & 10 & 3 & 4 \\
2 & 15 & 3 & 3 \\
3 & 20 & 2 & 4 \\
4 & 21 & 3 & 3 \\
5 & 50 & 4 & 4 \\
6 & 100 & 5 & 4 \\
7 & 200 & 6 & 5 \\
\hline
\end{tabular}
\label{tabdataset}
\end{center}
\end{table}

The datasets consist of non-symmetric sociometric matrices and requirements matrices, generated randomly. In order to illustrate the MTFP, the input matrices of database 1 are presented in detail in Tables \ref{tabreqmat} and \ref{tabsocmat}. The rest of the datasets and the source-code is available at:

\href{https://github.com/esgario/genetic-algorithm-to-mtfp}{https://github.com/esgario/genetic-algorithm-to-mtfp}

\begin{table}[!htbp]
\caption{Dataset 1 - Requirement Matrix}
\begin{center}
\begin{tabular}{c|ccc}
& $G_1$ & $G_2$ & $G_3$ \\ \hline
$D_1$ & 2 & 2 & 0 \\
$D_2$ & 2 & 1 & 0 \\
$D_3$ & 0 & 1 & 1 \\
$D_4$ & 0 & 0 & 1
\end{tabular}
\label{tabreqmat}
\end{center}
\end{table}

\begin{table}[!htbp]
\caption{Dataset 1 - Sociometric Matrix}
\begin{center}
\begin{tabular}{c|cccccccccc}
& $I_1$ & $I_2$ & $I_3$ & $I_4$ & $I_5$ & $I_6$ & $I_7$ & $I_8$ & $I_9$ & $I_{10}$ \\
\hline
$I_1$ & 0 & 1 & 0  & 0  & 1  & -1 & 1 & 1  & 1  & -1 \\
$I_2$ & 0 & 0 & 0  & 0  & 1  & 1  & 1 & 0  & -1 & 1  \\
$I_3$ & 1 & 1 & 0  & 1  & -1 & 1  & 1 & -1 & 1  & 1  \\
$I_4$ & 1 & 1 & 1  & 0  & 0  & 1  & 1 & 1  & 1  & 1  \\
$I_5$ & 0 & 0 & -1 & -1 & 0  & 1  & 1 & 0  & 0  & 0  \\
$I_6$ & 0 & 1 & 1  & 0  & 0  & 0  & 1 & -1 & 0  & 1  \\
$I_7$ & 1 & 1 & 0  & 0  & 0  & 0  & 0 & 1  & 1  & 0  \\
$I_8$ & 0 & 0 & 1  & 0  & 0  & 0  & 0 & 0  & 1  & 1  \\
$I_9$ & 1 & 0 & 0  & 0  & 0  & 0  & 0 & 0  & 0  & 0  \\
$I_{10}$ & 0 & 1 & -1 & 0  & 0  & 1  & 1 & 0  & -1 & 0 
\end{tabular}
\label{tabsocmat}
\end{center}
\end{table}

\subsection{Experimental Setup}

Experiments were performed in the development environment Matlab 2015a, running on a PC with Intel Core i5 processor, 8 GB of memory and Microsoft Windows 10 operating system.

Before performing the experiments it is necessary to set up values for GA parameters: number of generations $n_{gen}$, population size $n_p$, crossover probability $\beta$ and mutation probability $\alpha$. The effectiveness of the algorithm greatly depends on the parameters choices. So, empirical tests were performed seeking more appropriate values combination to this problem.

Aiming to estimate the number of generations it was taken into account that a Solution Matrix generated by a chromosome has dimension $n_i \times n_g$ in such a way that satisfying the constraint \eqref{eq:restricao2} the total combinations of possible solutions is $n_g^{n_i}$. In order to make the number of generations scalable for each problem, the information of the number of possible combinations was used so that $n_{gen} = n_g^{Kn_i}$ where $K$ is a constant. However, in this way the number of generations grows abruptly as the number of individuals increases. To overcome this problem a logarithmic transformation was performed, resulting in $n_{gen} = Kn_i\log{n_g}$. In the experiments performed $K=20$ was chosen.

For population size with $n_p > 50$, little improvements have been observed. Therefore, for all experiments was used $n_p=50$. The crossover probability did not have a significant impact on the results. In the experiments it was set up to $\beta=0.2$ which is equivalent to $\beta=0.8$. In the first case, approximately $20\%$ of the genes would be exchanged and in the second case $80\%$ of the genes would be exchanged, resulting in similar offspring pairs.

Initial tests showed that a fixed value of mutation probability for all datasets did not present good results. This happens due to the difference in the number of individuals of the problem at hand. In addition, the tests have shown that applying the mutation on average in only one gene of the chromosome provides good results. Therefore, since a chromosome has dimension equivalent to the number of individuals $n_i$ of the MTFP, it was set up to $\alpha=\frac{1}{n_i}$. Taking as an example a dataset with $n_i=50$ then we have $\alpha=0.02$, i.e., each gene has a $2\%$ chance of mutation.

Table \ref{tabparam} presents the parameters used in computational experiments with GA.

\begin{table}[htbp]
\caption{Parameters of Genetic Algorithm}
\begin{center}
\begin{tabular}{cc}
\hline
\textbf{\textit{Parameter}} & \textbf{\textit{Value}} \\
\hline
Number of generations & $n_{gen} = Kn_i\log{n_g}$ \\
Population size & $n_p = 50$\\
Crossover probability & $\beta = 0.2$ \\
Mutation probability & $\alpha=\frac{1}{n_i}$ \\
\hline
\end{tabular}
\label{tabparam}
\end{center}
\end{table}

\subsection{Results Analysis}

GA was performed on all datasets and statistical results were collected for $20$ runs and presented in the table \ref{tabresults}. In addition to the GA results, table \ref{tabresults} presents the results for the exhaustive method that generates all possible permutations satisfying the Requirement Matrix \eqref{eq:reqmatrix} and evaluates the cohesion for all permutations that meet the problem requirements. The exhaustive method results represent the optimal solution of the problem. For both algorithms, time was computed as the mean of $20$ runs.

\begin{table*}[htbp]
\caption{Simulation results for Exhaustive Method and Genetic Algorithm}
\begin{center}
\begin{tabular}{c|cccc|cccccc}
\hline
     & \multicolumn{4}{c}{\textbf{Exhaustive Method}}  & \multicolumn{6}{c}{\textbf{Genetic Algorithm}} \\
\cline{2-11} 
Dataset & Best Fitness & Time (s) & Permutations & Func. Eval. & Max & Mean & Std & Min & Time (s) & Func. Eval. \\
\hline
1 & 1.6000 & 1.8e-2 & 1296 & 36 & 1.6000 & 1.5800 & 0.0616 & 1.4000 & 1.2 & 11000 \\
\hline
2 & 2.3333 & 134.5 & 1.4e+7 & 7200 & 2.3333 & 2.2100 & 0.1087 & 2.0000 & 2.2 & 16500 \\
\hline
3 & 3.5000 & 234.6 & 2.5e+7 & 5040 & 3.5000 & 3.4425 & 0.0494 & 3.3500 & 2.0 & 13850 \\
\hline
4 & 2.6667 & 6082.9 & 6.1e+8 & 31360 & 2.6667 & 2.5667 & 0.1155 & 2.3810 & 4.0 & 23050 \\
\hline
5 & N/A & N/A & N/A & N/A & 3.1400 & 2.6600 & 0.2569 & 2.2000 & 37.5 & 69300 \\
\hline
6 & N/A & N/A & N/A & N/A & 4.1600 & 3.5350 & 0.2573 & 3.1400 & 289.7 & 160950  \\
\hline
7 & N/A & N/A & N/A & N/A & 5.5600 & 4.5930 & 0.4232 & 3.8500 & 2659.3 & 358350 \\
\hline
\end{tabular}
\label{tabresults}
\end{center}
\end{table*}

Due to the NP-Hard nature of the problem, the application of the exhaustive method was limited only to the less complex datasets. Among the first four datasets the only one whose exhaustive method presented the shortest time was the dataset 1, whose optimal solution is obtained with a small number of permutations. The Solution Matrix obtained for this dataset with the exhaustive method is presented in table \ref{tabsolmat}.

\begin{table}[!htbp]
\begin{center}
\caption{Dataset 1 - Solution Matrix}
\begin{tabular}{c|ccc}
& $G_1$ & $G_2$ & $G_3$ \\ \hline
$I_1$ & 1 &	0 &	0 \\
$I_2$ & 1 &	0 &	0 \\
$I_3$ & 0 &	1 &	0 \\
$I_4$ & 0 &	1 &	0 \\
$I_5$ & 1 &	0 &	0 \\
$I_6$ & 0 &	1 &	0 \\
$I_7$ & 1 &	0 &	0 \\
$I_8$ & 0 &	0 &	1 \\
$I_9$ & 0 &	1 &	0 \\
$I_{10}$ & 0 &	0 &	1
\end{tabular}
\label{tabsolmat}
\end{center}
\end{table}

Comparing both methods, it is clear how effective the GA is, especially when solving datasets 1-4 whose results is close enough to the optimal results obtained by the exhaustive method with a much shorter time at bases 2-4. The penalty method was quite effective for this problem, since GA was able to find feasible solutions in all datasets and runs. Although the optimal values obtained for the databases 5-7 are not known, $n_{gen}$ was adapted in such a way that for more complex problems the algorithm was run with a greater number of generations. The higher the value of $n_{gen}$ the greater the number of function evaluations, which leads the algorithm to explore more the search space.

\begin{figure}[htbp]
\centerline{\includegraphics[width=3.2in]{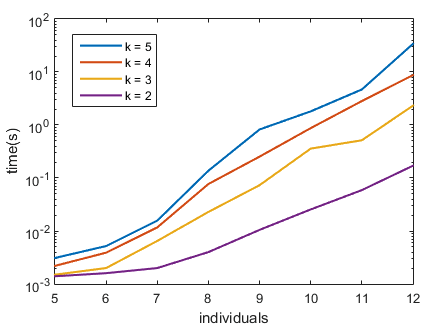}}
\caption{Average time to execute the exhaustive method by the number of individuals.}
\label{figtimeexec}
\end{figure}

In order to verify how restrictive is the MTFP due to its combinatorial nature, a series of experiments were performed in different scenarios (different input matrices) for the set of individuals $n_i = [5,...,12]$, for the groups $k = [2,...,5]$ and a fixed number of departments $j = 3$. Due to differences in input Requirement Matrices, the number of permutations performed by the exhaustive method is also different. In order to compute the average time, $20$ runs were performed for each value of $n_i$ and $k$. The discrepancy values (outliers) of maximum and minimum were discarded and the mean was calculated from the remaining values. The average execution time of the exhaustive method is shown in Fig. \ref{figtimeexec}. The similarity of the mean time curves obtained with a straight line in the semi-log plot, suggests an exponential growth of time when the number of individuals increases.

\section{Conclusion}

In this paper, an evolutionary approach was proposed to solve the MTFP based on sociometry in order to form teams with high cohesion in a reduced time. The genetic operators were adapted to deal with the constraints of the problem. In addition, the penalty method was used to force the algorithm to find feasible solutions. Computational experiments were performed to evaluate the performance of the proposed algorithm. The algorithm was executed for datasets with different levels of complexity and the statistical results of cohesion and time are presented. The performance of the algorithm was compared with the exhaustive method in four out of seven datasets. The comparison with the rest of the datasets was impracticable due to the high computational time required by the exhaustive method for more complex problems. The proposed method provides results close to the optimal one with a reduced computational time. So, the approach turns out to be quite promising. Future research will investigate the use of the matrix of allocation with fractional values which will allow the distribution of the workload of employees in more than one group.

\section*{Acknowledgment}

This study was financed in part by the Coordena\c{c}\~ao de Aperfei\c{c}oamento de Pessoal de N\'ivel Superior - Brasil (CAPES) - Finance Code 001. R. A. Krohling would like to thank the Brazilian agency CNPq and the local Agency of the state of Esp{\'i}rito Santo FAPES for financial support under grant No. 309161/2015-0 and No. 039/2016, respectively.


\ifCLASSOPTIONcaptionsoff
  \newpage
\fi

\bibliographystyle{ieeetr}


\end{document}